# Fast Bounded Online Gradient Descent Algorithms for Scalable Kernel-Based Online Learning

Peilin Zhao[†]　　　　　　　　　　　　　　　　　　　　ZHAO0106@NTU.EDU.SG
Jialei Wang[†]　　　　　　　　　　　　　　　　　　　　JL.WANG@NTU.EDU.SG
Pengcheng Wu[†]　　　　　　　　　　　　　　　　　　WUPE0003@NTU.EDU.SG
Rong Jin[‡]　　　　　　　　　　　　　　　　　　　　　RONGJIN@CSE.MSU.EDU
Steven C.H. Hoi[†]　　　　　　　　　　　　　　　　　　CHHOI@NTU.EDU.SG

[†]School of Computer Engineering, Nanyang Technological University, Singapore 639798
[‡]Department of Computer Science and Engineering, Michigan State University, USA

## Abstract

Kernel-based online learning has often shown state-of-the-art performance for many online learning tasks. It, however, suffers from a major shortcoming, that is, the unbounded number of support vectors, making it non-scalable and unsuitable for applications with large-scale datasets. In this work, we study the problem of bounded kernel-based online learning that aims to constrain the number of support vectors by a predefined budget. Although several algorithms have been proposed in literature, they are neither computationally efficient due to their intensive budget maintenance strategy nor effective due to the use of simple Perceptron algorithm. To overcome these limitations, we propose a framework for bounded kernel-based online learning based on an online gradient descent approach. We propose two efficient algorithms of bounded online gradient descent (BOGD) for scalable kernel-based online learning: (i) BOGD by maintaining support vectors using uniform sampling, and (ii) BOGD++ by maintaining support vectors using non-uniform sampling. We present theoretical analysis of regret bound for both algorithms, and found promising empirical performance in terms of both efficacy and efficiency by comparing them to several well-known algorithms for bounded kernel-based online learning on large-scale datasets.

## 1. Introduction

The goal of kernel-based online learning is to sequentially update a nonlinear kernel-based classifier given a sequence of training examples (Kivinen et al., 2001; Cheng et al., 2006; Crammer et al., 2006; Jin et al., 2010; Zhao et al., 2011). Although it yields significantly better performance than linear online learning, the main shortcoming of kernel-based online learning is its potentially unbounded number of support vectors, which requires a large amount of memory for storing support vectors and a high computational cost per iteration, both making it unsuitable for large-scale applications. In this work, we address this challenge by developing a computationally efficient framework for budget online learning in which the number of support vectors is bounded by a predefined size (i.e., budget).

In literature, several algorithms have been proposed for online budget learning. Crammer et al. (Crammer et al., 2003) proposed a heuristic approach for online budget learning, which was further improved in (Weston & Bordes, 2005). The basic idea of these two algorithms is to remove the support vector that has the least impact on the classification performance when the budget number of support vectors is reached. The main shortcoming of these two algorithms is that they are heuristic approaches and do not have solid theoretic supports (i.e., neither a mistake bound nor a regret bound is proved).

Forgetron (Dekel et al., 2005) is the first online budget learning algorithm that has guarantee on the number of mistakes. At each iteration, if the classifier makes a mistake, it conducts a three-step updating: it first runs the standard Perceptron (Rosenblatt, 1958) updating; it then shrinks the weights of support vectors by a carefully chosen scaling factor; it finally removes the support vector with the least weight. Randomized Budget Perceptron (RBP) (Cavallanti et al., 2007) removes a randomly selected support vector when the number of support vectors exceeds the predefined budget. It achieves similar mistake bound and empirical performance as the Forgetron algorithm.

Unlike the strategy that discards one of support vectors to maintain the budget, Projectron (Orabona et al., 2008) adopts a projection strategy to bound the number of support vectors. Specifically, in each iteration when the training example is misclassified, it first constructs a new kernel classifier by applying the updating rule of Perceptron to the current classifier; it then projects the new classifier into the space spanned by all the support vectors except the new example received. The classifier will remain unchanged if the difference between the new classifier and its projection is smaller than a given threshold. Empirical studies show that Projectron usually outperforms Forgetron in classification but with significantly longer running time. One main shortcoming of Projectron is that although the number of support vectors of Projectron is bounded, it is however unclear the exact number of support vectors achieved by Projectron in theory. In addition, its high computational cost makes it unsuitable for large-scale applications.

All the existing algorithms for online budget learning are based on the Perceptron algorithm, partially because they are mostly concerned with the mistake bound, not the regret bound. In this paper, we develop a "Bounded Online Gradient Descent" (BOGD) framework for online budget learning algorithms, based on the online gradient descent algorithms (Kivinen et al., 2001; Zinkevich, 2003; Ying & Pontil, 2008). Similar to the Random Budget Perceptron, the proposed algorithms randomly select one of the existing support vectors to discard when the buffer of support vectors overflows. However, unlike the Random Budget Perceptron that discards every support vector with equal probability, in one of our algorithms, the probability of discarding a support vector depends on its weight, making it more effective for online budget learning. Different from most existing studies that can only obtain a guarantee on the mistake bound, we derive regret bounds for the proposed algorithms, making it possible to convert the proposed algorithms into batch learning algorithms when the received examples are iid samples. Finally, it is important to distinguish the proposed work from sparse online learning (Langford et al., 2009; Duchi & Singer, 2009) whose goal is to learn a sparse *linear* classifier from a sequence of training examples. In contrast, we focus on learning a nonlinear kernel classifier.

The rest of the paper is organized as follows. Section 2 introduces the basic setting of online budget learning, and presents both theoretical and algorithmic details of the proposed approaches for online budget learning. Section 3 discusses our empirical studies on six real world datasets. Section 4 concludes this work.

## 2. Algorithms and Analysis

We consider kernel-based online learning for classification. Our goal is to learn a function $f : \mathbb{R}^d \to \mathbb{R}$ from a sequence of training examples $\{(\mathbf{x}_t, y_t), t \in [T]\}$, where $\mathbf{x}_t \in \mathbb{R}^d$, $y_t \in \mathcal{Y} = \{-1, +1\}$ and $[T] = \{1, \ldots, T\}$. We predict the class assignment for $\mathbf{x}$ by $\text{sgn}(f(\mathbf{x}))$, and measure the classification confidence by $|f(x)|$. Let $\ell(yf(\mathbf{x})) : \mathbb{R} \to \mathbb{R}$ be a convex loss function that is Lipschitz continuous with Lipschitz constant $L$. Let $\mathcal{H}$ be an RKHS endowed with a kernel function $\kappa(\cdot, \cdot) : \mathbb{R}^d \times \mathbb{R}^d \to \mathbb{R}$. We assume $\kappa(\mathbf{x}, \mathbf{x}) \leq 1$ for any $\mathbf{x} \in \mathbb{R}^d$. Similar to kernel-based online learning (Kivinen et al., 2001; Zinkevich, 2003) and the Pegasos algorithm (Shalev-Shwartz et al., 2011), at each trial of online learning, given a received training example $(\mathbf{x}_t, y_t)$, we define the following loss function:

$$\mathcal{L}_t(f) = \frac{\lambda}{2}\|f\|_\mathcal{H}^2 + \ell(y_t f(\mathbf{x}_t)) \quad (1)$$

We first describe an online learning algorithm, similar to kernel-based online learning (Kivinen et al., 2001; Zinkevich, 2003), that minimizes the regret of $\sum_{t=1}^{T} \ell_t(f_t)$ using the online gradient descent approach. At each trial $t$, given the classifier $f_t$ and training example $(\mathbf{x}_t, y_t)$, we update the classifier by

$$\begin{aligned}f_{t+1}(\cdot) &= f_t(\cdot) - \eta\nabla\mathcal{L}_t(f_t) \\ &= (1-\eta\lambda)f_t(\cdot) - \eta y_t \ell'(y_t f_t(\mathbf{x}_t))\kappa(\mathbf{x}_t, \cdot)\end{aligned} \quad (2)$$

where $\eta$ is the stepsize and $\lambda > 0$ is the regularization parameter.

**Theorem 1.** *Let $f_t, t \in [T]$ be a sequence of classifiers generated by the updating rule in (2). We have the following bound for any $f \in \mathcal{H}$,*

$$\sum_{t=1}^{T} \ell(y_t f_t(\mathbf{x}_t)) \leq \frac{\lambda T + \eta^{-1}}{2}\|f\|_\mathcal{H}^2 + \sum_{t=1}^{T} \ell(y_t f(\mathbf{x}_t)) + \eta L^2 T$$

By setting $\lambda T = \eta^{-1}$, we have

$$\sum_{t=1}^{T} \ell(y_t f_t(\mathbf{x}_t)) \leq \frac{1}{\eta}\|f\|_\mathcal{H}^2 + \sum_{t=1}^{T} \ell(y_t f(\mathbf{x}_t)) + \eta L^2 T$$

which leads to $O(1/\sqrt{T})$ bound if $\eta = O(1/\sqrt{T})$. Note that we did not exploit the strong convexity of $\mathcal{L}_t(f)$,

which often leads to a better bound. This is because our goal is to bound $\sum_t \ell_t(y_t f_t(\mathbf{x}_t))$, not $\sum_t \mathcal{L}_t(f_t)$. In addition, to exploit the strong convexity of $\mathcal{L}_t(f)$, we have to vary the stepsize $\eta$ over trials, making it difficult to extend the analysis to online budget learning.

We now modify the updating rule in (2) for online budget learning. The first modification is to introduce a domain to which the updated classifier will be projected. Specifically, we define the domain $\Omega$ as:

$$\Omega(\eta\gamma) = \left\{ f(\cdot) = \sum_{t=1}^{T} \alpha_t y_t \kappa(\mathbf{x}_t, \cdot) : \alpha_t \in [0, \gamma\eta], t \in [T] \right\} \quad (3)$$

where $\gamma\eta > 0$ specifies the maximum weight that can be assigned to any support vector. Using the domain $\Omega(\eta\gamma)$, we modify the updating rule in (2) as follows

$$f_{t+1} = \pi_{\Omega(\eta\gamma)}(f_t - \eta \nabla \mathcal{L}_t(f_t)) \quad (4)$$

where $\pi_{\Omega(\eta\gamma)}(f)$ projects $f$ into the domain $\Omega(\eta\gamma)$. Note that when $\gamma \geq L$, we have $\pi_{\Omega(\eta\gamma)}(f) = f$ because the weights for support vectors never increase over trials and for any support vector, its initially assigned weight is $\eta L$.

Let $B > 0$ be a predefine budget. Our goal is to bound the number of support vector by $B$. When the number of the support vectors in $f(\cdot)$ is less than $B$, we simply run the updating rule in (4) without any change. Without loss of generality, we consider a trial $t$ where the number of support vectors in $f_t(\cdot)$ is $B$ and we need to update $f_t(\cdot)$ with a new training example $(\mathbf{x}_t, y_t)$. Note that the gradient of $\mathcal{L}_t(f_t)$ is written as $\lambda f_t(\cdot) + y_t \ell'(y_t f_t(\mathbf{x}_t))\kappa(\mathbf{x}_t, \cdot)$. Our strategy is to approximate $f_t(\cdot)$ in $\nabla \mathcal{L}(f)$ with its unbiased estimator $\widehat{f}_t(\cdot)$ so that the updated classifier $f_{t+1}(\cdot) = f_t - \eta\lambda \widehat{f}_t - \eta y_t \ell'(y_t f_t(\mathbf{x}_t))\kappa(\mathbf{x}_t, \cdot)$ contains exactly $B$ support vectors. More specifically, we express the classifier $f_t(\cdot)$ as

$$f_t(\cdot) = \sum_{i=1}^{B} \alpha_i^t y_i^t \kappa(\mathbf{x}_i^t, \cdot)$$

where $\{(\mathbf{x}_i^t, y_i^t), i \in [B]\}$ are the support vectors and $\alpha_i^t > 0$ is the weight for support vector $(\mathbf{x}_i^t, y_i^t)$. In order to generate an unbiased estimator $\widehat{f}_t(\cdot)$ for $f_t(\cdot)$, we randomly select one support vector according to a distribution $\mathbf{p}^t = (p_1^t, \ldots, p_B^t)$. We introduce a binary variable $Z_i^t$, with $Z_i^t = 1$ indicating that support vector $(\mathbf{x}_i^t, y_i^t)$ is selected and zero otherwise. Evidently, $\sum_{i=1}^{B} Z_i^t = 1$. Based on $Z^t = (Z_1^t, \ldots, Z_B^t)$, we consider the following general form for constructing the unbiased estimator $\widehat{f}_t(\cdot)$

$$\widehat{f}_t(\cdot) = \sum_{i=1}^{B} \left( a_i^t Z_i^t + b_i^t \right) y_i^t \kappa(\mathbf{x}_i^t, \cdot) \quad (5)$$

where $a_i^t \geq 0$ and $b_i^t$ are parameters that need to be determined. To ensure $\mathrm{E}[\widehat{f}_t(\cdot)] = f_t(\cdot)$, we have the following condition for $a_i^t$ and $b_i^t$

$$a_i^t p_i^t + b_i^t = \alpha_i^t, i \in [B] \quad (6)$$

Using the unbiased estimator $\widehat{f}_t(\cdot)$, we have the classifier $f_t(\cdot)$ updated as

$$\begin{aligned} f_{t+1}(\cdot) &= \pi_{\Omega(\eta\gamma)}\Big( -\eta \ell'(y_t f_t(\mathbf{x}_t)) y_t \kappa(\mathbf{x}_t, \cdot) \\ &\quad + \sum_{i=1}^{B} \left( \alpha_i^t - \lambda\eta[b_i^t + a_i^t Z_i^t] \right) y_i^t \kappa(\mathbf{x}_i^t, \cdot) \Big) \end{aligned} \quad (7)$$

In order to ensure that the number of support vectors in $f_{t+1}(\cdot)$ is $B$, we need to have one of the coefficients in (7) set to zero, leading to the following condition for $a_i^t$ and $b_i^t$.

$$\alpha_i^t = \lambda\eta(b_i^t + a_i^t), \ i \in [B] \quad (8)$$

Conditions (6) and (8) are the key for designing the sampling probabilities $\mathbf{p}^t$ and weights $(a_i^t, b_i^t)$ for each support vector. Given $\mathbf{p}_t$, we have the following expression for $(a_i^t, b_i^t)$

$$a_i^t = \frac{1 - \lambda\eta}{\lambda\eta(1 - p_i^t)} \alpha_i^t, \quad b_i^t = \frac{\lambda\eta - p_i^t}{\lambda\eta(1 - p_i^t)} \alpha_i^t, \quad i \in [B] \quad (9)$$

As a result, the weight $\alpha_i^{t+1}$ is updated as follows

$$\alpha_i^{t+1} = \min\left( (1 - Z_i^t)\frac{1 - \lambda\eta}{1 - p_i^t} \alpha_i^t, \gamma\eta \right), \ i \in [B] \quad (10)$$

According to (10), the weight for the selected support vector (i.e., $Z_t^i = 1$) is set to zero in the updated classifier $f_{t+1}(\cdot)$, implying that the selected support vector is discarded from the updated classifier. Finally, Algorithm 1 summarizes the proposed framework of Bounded Online Gradient Descent (BOGD) learning.

Given the sampling probabilities $\mathbf{p}_t, t \in [T]$, we have the following theorem for Algorithm 1.

**Theorem 2.** *Assume $\kappa(\mathbf{x}, \mathbf{x}) \leq 1$ and $\lambda\eta \leq 1/2$. Let $f_t, t \in [T]$ be the sequence of classifiers generated by Algorithm 1. Then, for any $f \in \Omega(\eta\gamma)$, we have in expectation the overall loss bounded as follows*

$$\mathrm{E}\left[\sum_{t=1}^{T} \ell(y_t f_t(\mathbf{x}_t))\right] \leq \frac{\eta^{-1} + \lambda T}{2} \|f\|_{\mathcal{H}}^2 + \sum_{t=1}^{T} \ell(y_t f(\mathbf{x}_t))$$

$$+ \eta L^2 T + \frac{(1 - \lambda\eta)^2}{\eta} \mathrm{E}\left[\sum_{t \in V_T} \sum_{i=1}^{B} \frac{p_i^t [\alpha_i^t]^2}{(1 - p_i^t)^2}\right]$$

*where $V_T = [T]/U_T$ and $U_T = \{t \in [T] \mid |SV_t| < B \text{ or } \ell'(y_t f_t(\mathbf{x}_t)) = 0\}$.*

**Algorithm 1** A framework of Bounded Online Gradient Descent (BOGD)

**Input**: the maximum budget size $B$, stepsize $\eta$, regularization parameter $\lambda > 0$, and maximum coefficient $\gamma > 0$.
**Initialize** $\mathcal{S}_1 = \emptyset$, $f_1 = 0$.
**for** $t = 1, 2, \ldots, T$ **do**
   Receive $\mathbf{x}_t$;
   Predict $\widehat{y}_t = \mathrm{sgn}(f_t(\mathbf{x}_t))$;
   Receive $y_t$ and suffer loss $\ell(y_t f_t(\mathbf{x}_t))$;
   **if** $\ell'(y_t f_t(\mathbf{x}_t)) = 0$ **then**
     $f_{t+1}(\cdot) = (1 - \eta\lambda)f_t(\cdot)$ and $\mathcal{S}_{t+1} = \mathcal{S}_t$.
   **else**
     **if** $|\mathcal{S}_t| < B$ **then**
       $f_{t+1}(\cdot) = (1-\eta\lambda)f_t(\cdot) - \eta\ell'(y_t f_t(\mathbf{x}_t))y_t\kappa(x_t,\cdot)$
       and $\mathcal{S}_{t+1} = \mathcal{S}_t \cup \{t\}$.
     **else**
       Compute the sampling distribution $\mathbf{p}_t = (p_1^t, \ldots, p_B^t)$;
       Sample an index $i_k$ from $\{1, \ldots, B\}$ according to distribution $\mathbf{p}_t$;
       Set $Z_{i_k}^t = 1$ and $Z_i^t = 0$, $i \in [B]/\{i_k\}$;
       $a_i^t = \frac{1-\lambda\eta}{\lambda\eta(1-p_i^t)}\alpha_i^t$, $b_i^t = \frac{\lambda\eta - p_i^t}{\lambda\eta(1-p_i^t)}\alpha_i^t$, $i = 1, \ldots, B$;
       $f_{t+1}(\cdot) = $ Eq. (7);
       $\mathcal{S}_{t+1} = \mathcal{S}_t \cup \{t\}/\{i_k\}$.
     **end if**
   **end if**
**end for**

*Proof.* Using the standard analysis of gradient descent (Kivinen et al., 2001; Zinkevich, 2003), it is not difficult to show for any $f \in \Omega(\eta\gamma)$,

$$\mathrm{E}\left[\sum_{t=1}^T \left\{\frac{\lambda}{2}\|f_t\|_{\mathcal{H}}^2 + \ell(y_t f_t(\mathbf{x}_t))\right\}\right] - \sum_{t=1}^T \left\{\frac{\lambda}{2}\|f\|_{\mathcal{H}}^2 + \ell(y_t f(\mathbf{x}_t))\right\}$$
$$\leq \frac{\|f\|_{\mathcal{H}}^2}{2\eta} + \eta L^2 T + \eta \mathrm{E}\left[\sum_{t=1}^T \lambda^2 \|\widehat{f}_t\|_{\mathcal{H}}^2\right] \quad (11)$$

We consider two scenarios:

**Case 1**: Consider the trial $t \in U_T$. Since no sampling is done in these trials, we thus have $\mathrm{E}_t[|\widehat{f}_t|_{\mathcal{H}}^2] = \|f_t\|_{\mathcal{H}}^2$.
**Case 2**: Consider the trial $t \in V_T$, we have

$\mathrm{E}_t[|\widehat{f}_t|_{\mathcal{H}}^2]$
$= \|f_t\|_{\mathcal{H}}^2 - \left\|\sum_{i=1}^B a_i^t p_i^t y_i^t \kappa(\mathbf{x}_i^t, \cdot)\right\|_{\mathcal{H}}^2 + \sum_{i=1}^B p_i^t [a_i^t]^2 \kappa(\mathbf{x}_i^t, \mathbf{x}_i^t)$
$\leq \|f_t\|_{\mathcal{H}}^2 + ([\lambda\eta]^{-1} - 1)^2 \sum_{i=1}^B \frac{p_i^t [\alpha_i^t]^2}{(1-p_i^t)^2}$

We complete the proof by substituting into (11) the above expression for $\|f\|_{\mathcal{H}}^2$. □

Below, we discuss two different designs of sampling probabilities $\mathbf{p}_t$, i.e., (i) uniform sampling, and (ii) non-uniform sampling.

**Uniform Sampling.** In this approach, we set $p_i^t = 1/B$ for any $i \in [B]$ and any $t \in [T]$. According to Theorem 2, it is not difficult to have the following result for the loss bound.

**Theorem 3.** *For any classifier $f \in \Omega(\eta\gamma)$, we have the following bound for Algorithm 1 using uniform sampling*

$$\mathrm{E}\left[\sum_{t=1}^T \ell(y_t f_t(\mathbf{x}_t))\right] \leq \left(\left(\frac{B}{B-1}\right)^2 \gamma^2 + L^2\right)\eta T$$
$$+ \frac{\eta^{-1} + \lambda T}{2}\|f\|_{\mathcal{H}}^2 + \sum_{t=1}^T \ell(y_t f(\mathbf{x}_t)) = A(\eta) + C(\eta) \quad (12)$$

*where*

$$A(\eta) = \left(\left(\frac{B}{B-1}\right)^2 \gamma^2 + L^2\right)\eta T + \frac{\eta^{-1} + \lambda T}{2}\|f\|_{\mathcal{H}}^2,$$
$$C(\eta) = \sum_{t=1}^T \ell(y_t f(\mathbf{x}_t)).$$

*Let $\lambda\eta T = 1$ and $\eta = 1/\sqrt{T}$. We then have, for any $f \in \Omega(\eta\gamma)$, that*

$$\mathrm{E}\left[\sum_{t=1}^T \ell(y_t f_t(\mathbf{x}_t))\right] - \sum_{t=1}^T \ell(y_t f(\mathbf{x}_t))$$
$$\leq \left(\left(\frac{B}{B-1}\right)^2 \gamma^2 + L^2\right)\sqrt{T} + \|f\|_{\mathcal{H}}^2 \sqrt{T} = O(\sqrt{T}) \quad (13)$$

**Remark.** We have two comments for the above results. First, by choosing different stepsize $\eta$, we make a tradeoff between $A(\eta)$ and $C(\eta)$. In particular, a small $\eta$ will result in a small value for $A(\eta)$ but a large value for $C(\eta)$. This is because a small $\eta$ reduces the size of hypothesis space $\Omega(\eta\gamma)$ and consequentially increases the overall loss $\sum_{t=1}^T \ell(y_t f(\mathbf{x}_t))$. Similarly, a large $\eta$ will lead to large $A(\eta)$ but potentially small $C(\eta)$. Second, although (13) shows a regret bound of $O(\sqrt{T})$ independent from $B$, it does not contradict the analysis presented in (Dekel et al., 2005). This is because we restrict the competitor $f$ to the domain $\Omega(\eta\gamma)$ while the analysis in (Dekel et al., 2005) considers any hypothesis in RHKS $\mathcal{H}$ as a competitor. Observe that the projection $\pi_{\Omega(\eta\gamma)}(f)$ in (7) is no longer in effect if we set $\gamma \geq L$ and $\lambda\eta \geq 1/B$ in our algorithm. As a result, under the conditions $\gamma \geq L$ and $\lambda\eta \geq 1/B$, for any classifier $f \in \mathcal{H}$, with appropriate choice of $\eta$ and $\lambda$, we have the following regret bound for the the sequence of classifier generated by Algorithm 1 using uniform sampling:

$$\mathrm{E}\left[\sum_{t=1}^T \ell(y_t f_t(\mathbf{x}_t))\right] - \sum_{t=1}^T \ell(y_t f(\mathbf{x}_t)) \leq O\left(\frac{T}{\sqrt{B}}\|f\|_\mathcal{H}\right) \quad (14)$$

As indicated by the regret bound in (14), if we consider any $f \in \mathcal{H}$ as a competitor, unless we set $B = T$, we will not be able to obtain an $O(\sqrt{T})$ regret bound. Although this result may seem significantly worse than the one presented (Dekel et al., 2005), we emphasize that (14) is about regret bound while the result in (Dekel et al., 2005) is about mistake bound. In general, deriving a good regret bound is usually more challenging than getting a similar mistake bound.

**Nonuniform Sampling.** To fully exploit the information we have about the classifier $f(\cdot) = \sum_{i=1}^B \alpha_i y_i \kappa(\mathbf{x}_i, \cdot)$, we consider a nonuniform sampling approach to BOGD by choosing the values of $\mathbf{p}$ as follows:

$$(1 - p_i)^2 = s^2 \alpha_i^2 \kappa(\mathbf{x}_i, \mathbf{x}_i), i \in [B] \quad (15)$$

where $s$ is the normalization factor. Given the expression in (15), it is straightforward to derive $p_i$ as follows

$$p_i = 1 - s\alpha_i\sqrt{\kappa(\mathbf{x}_i, \mathbf{x}_i)}, \quad (16)$$

where $s = \frac{(B-1)}{\sum_{i=1}^B \alpha_i \sqrt{\kappa(\mathbf{x}_i,\mathbf{x}_i)}}$.

Before presenting the regret bound, we define function $H(f)$ that measures the skewness of the coefficients for the support vectors used by $f$. More specifically, $H(f)$ is defined as

$$H(f) = B\sum_{i=1}^B \alpha_i^2 \kappa(x_i, x_i) - \left(\sum_{i=1}^B \alpha_i \sqrt{\kappa(x_i, x_i)}\right)^2$$

According to Cauchy-Schwarz inequality, we always have $H(f) \geq 0$ where the equality holds if and only if $\alpha_1 = \ldots = \alpha_B$.

**Theorem 4.** *Assume $\kappa(\mathbf{x}, \mathbf{x}) \leq 1$ and $\kappa(\mathbf{x}, \mathbf{x}') \geq 0$. Let $f_t, t \in [T]$ be the sequence of classifiers generated by Algorithm 1 using the nonuniform sampling specified in (16). Then, for any $f \in \Omega(\eta\gamma)$, we have in expectation the loss experienced by $\{f_t\}_{t=1}^T$ bounded as:*

$$\mathrm{E}\left[\sum_{t=1}^T \ell(y_t f_t(\mathbf{x}_t))\right] - \sum_{t=1}^T \ell(y_t f(\mathbf{x}_t)) \leq$$

$$((\frac{B}{B-1})^2\gamma^2 + L^2)\eta T + \frac{\eta^{-1} + \lambda T}{2}\|f\|_\mathcal{H}^2 - \frac{(1-\lambda\eta)^2}{\eta(B-1)^2}\sum_{t \in V_t} H_t$$

*where $H_t = H(f_t)$ and $V_t$ is the set of trials where one of the support vector is discarded.*

*Proof.* The proof is almost identical to that of Theorem 2. The only difference is in bounding $\mathrm{E}_t[|\widehat{f}_t|_\mathcal{H}^2]$, $t \in V_t$, i.e.,

$\mathrm{E}_t[|\widehat{f}_t|_\mathcal{H}^2]$

$$\leq \|f_t\|_\mathcal{H}^2 + \left(\frac{1-\lambda\eta}{\lambda\eta}\right)^2 \frac{\left(\sum_{i=1}^B \alpha_i^t \sqrt{\kappa(\mathbf{x}_i^t, \mathbf{x}_i^t)}\right)^2}{(B-1)^2}$$

$$\leq \|f_t\|_\mathcal{H}^2 + \frac{(1-\lambda\eta)^2}{(\lambda\eta)^2(B-1)^2}\left(B\sum_{i=1}^B [\alpha_i^t]^2 \kappa(\mathbf{x}_i^t, \mathbf{x}_i^t) - H_t\right)$$

The rest of the proof follows almost the exactly same steps as that for Theorem 2. □

The above theorem clearly indicates that nonuniform sampling reduces the regret bound by taking advantage of the skewed distribution of coefficients assigned to support vectors. We note that although Theorem 4 does not give an explicit bound for the advantage of exploiting the skewed distribution of coefficients, it does show up significantly in our empirical study.

## 3. Experimental Results

In this section, we evaluate the empirical performance of the proposed algorithms for Bounded Online Gradient Descent (BOGD) learning algorithms by comparing them to the state-of-the-art algorithms for online budget learning.

### 3.1. Experimental Testbed

Table 1 shows the details of six binary-class datasets used in our experiments. All of these datasets can be downloaded from LIBSVM website [1] and UCI machine learning repository [2]. These datasets were chosen fairly randomly to cover a variety of datasets of different sizes.

Table 1. Details of the datasets in our experiments.

| Dataset | # instances | # features |
|---------|-------------|------------|
| german | 1000 | 24 |
| spambase | 4601 | 57 |
| magic04 | 19020 | 10 |
| w8a | 24692 | 300 |
| ijcnn1 | 141691 | 22 |
| codrna | 271617 | 8 |

---

[1] http://www.csie.ntu.edu.tw/~cjlin/libsvmtools/
[2] http://www.ics.uci.edu/~mlearn/MLRepository.html

Fast Bounded Online Gradient Descent Algorithms

| Algorithm | Perceptron | | | | OGD | | |
|---|---|---|---|---|---|---|---|
| Datasets | Mistake (%) | Support Vectors (#) | Time (s) | | Mistakes (%) | Support Vectors (#) | Time (s) |
| german | 34.805 ± 1.017 | 348.050 ± 10.175 | 0.069 | | 30.115 ± 0.618 | 583.550 ± 6.613 | 0.128 |
| spambase | 24.957 ± 0.460 | 1148.250 ± 21.166 | 0.486 | | 21.588 ± 0.303 | 2391.750 ± 13.973 | 1.071 |
| w8a | 3.501 ± 0.053 | 2264.950 ± 34.092 | 17.055 | | 2.343 ± 0.020 | 3352.900 ± 11.149 | 27.795 |
| magic04 | 27.093 ± 0.326 | 5153.100 ± 62.060 | 5.363 | | 20.176 ± 0.144 | 14333.350 ± 18.368 | 25.652 |
| ijcnn1 | 12.361 ± 0.120 | 17514.400 ± 169.788 | 438.169 | | 9.181 ± 0.030 | 25267.750 ± 39.840 | 640.728 |
| codrna | 14.038 ± 0.033 | 38128.800 ± 88.755 | 1392.621 | | 10.467 ± 0.024 | 51423.900 ± 74.865 | 1782.763 |

Table 2. Evaluation of non-budget algorithms on the the data sets.

### 3.2. Baseline Algorithms and Setup

We refer to as "BOGD" the proposed BOGD algorithm using uniform sampling, and as "BOGD++" the proposed BOGD algorithm using nonuniform sampling. We compare the two proposed BOGD algorithms with the following state-of-the-art algorithms for online budget learning: (i) "RBP" — the Random Budget Perceptron algorithm (Cavallanti et al., 2007), (ii) "Forgetron" — the Forgetron algorithm (Dekel et al., 2005), (iii) "Projectron" — the Projectron algorithm (Orabona et al., 2008), and (iv) "Projectron++" — the aggressive version of Projectron algorithm (Orabona et al., 2008). We also compare the proposed algorithms to two non-budget online learning algorithms: (i) "Perceptron" — the classical Perceptron algorithm (Rosenblatt, 1958), and (ii) "OGD" — the Online Gradient Descent algorithm (Kivinen et al., 2001; Ying & Pontil, 2008).

To make a fair comparison, all the algorithms in our comparison adopt the same experimental setup. The loss function $\ell$ is set as the hinge loss, i.e., $\ell(yf(\mathbf{x})) = \max(0, 1 - yf(\mathbf{x}))$. A Gaussian kernel is adopted in our study, for which the kernel width is set to 8 for all the algorithms and datasets. The regularization parameter $\lambda$, stepsize $\eta$ and parameter $\gamma$ in the proposed algorithm are selected using cross validation for all combinations of the datasets, algorithms and budgets(More specifically, $\lambda$ is chosen from $\{\frac{2^{-3}}{T^2}, \frac{2^{-2}}{T^2}, ..., \frac{2^3}{T^2}\}$ where $T$ is the number of instances; $\eta$ is chosen from $\{2^{-3}, 2^{-2}, ..., 2^3\}$; $\gamma$ is chosen from $\{2^0, 2^1, ..., 2^4\}$). The budget sizes $B$s for different datasets are set as proper fractions of the support vectors numbers of Perceptron, which are shown in Table 3. All the experiments were conducted 20 times, each with a different random permutation of data points. All the results were reported by averaging over the 20 runs. For performance metrics, we evaluate the online classification performance by mistake rates and running time. Finally, all of the algorithms were implemented in C++, and all experiments were run in a linux machine with 2.5GHz CPU.

### 3.3. Evaluation of Non-budget Algorithms

Table 2 summarizes the average performance of the two non-budget algorithms for kernel-based online learning. First, we find that OGD outperforms Perceptron significantly for all datasets according to t-test results, which implies that a budget OGD algorithm might be more effective than that based on the Perceptron algorithm. Second, the support vector size of OGD is in general much larger than that of Perceptron. Finally, the time cost of OGD is much higher than that of Perceptron, mostly due to the larger number of support vectors. Both the large number of support vectors and high computational time motivate the need of developing budget OGD algorithms.

### 3.4. Evaluation of Budget Algorithms

Table 3 summarizes the results of different budget online learning algorithms. First, we observe that RBP and Forgetron achieve similar performance for almost all cases. In addition, we also find that Projectron++ achieves a lower mistake rate than Projectron for almost all the datasets and for all budge sizes, which is similar to the results in (Orabona et al., 2008).

Second, compared to the baseline algorithms for online budget learning, the proposed BOGD algorithm achieves comparable, sometimes better mistake rates, especially when the budget size is large, demonstrating the effectiveness of our framework. Among all the compared algorithms for online budget learning, when the budget is large, we find that BOGD++ always achieves the lowest mistake rates for most the cases; when the budget is small, BOGD++ often achieves the best or close to the best results (except two datasets "german" and "spambase"). These results indicate the importance of exploiting the skewed distribution of weights for support vectors. Moreover, it is interesting to find that on most datasets (e.g., "german", "w8a", "ijcnn1", "magic04" and "codrna"), BOGD++ can even achieve significantly lower mistake rates than Perceptron that does not a budget constraint.

Fast Bounded Online Gradient Descent Algorithms| Budget Size | | B=100 | | B=150 | | B=200 | |
|---|---|---|---|---|---|---|---|
| Dataset | Algorithm | Mistake (%) | Time (s) | Mistakes (%) | Time (s) | Mistakes (%) | Time (s) |
| german | RBP | 38.060 %± 1.254 | 0.032 | 37.040 %± 0.658 | 0.044 | 35.740 %± 1.566 | 0.056 |
| | Forgetron | 37.320 %± 1.040 | 0.037 | 36.780 %± 1.894 | 0.041 | 36.280 %± 0.726 | 0.038 |
| | Projectron | 35.240 %± 0.635 | 0.041 | 35.100 %± 0.539 | 0.062 | 34.960 %± 0.853 | 0.097 |
| | Projectron++ | 34.500 %± 1.355 | 0.059 | 35.240 %± 0.814 | 0.109 | 34.600 %± 0.671 | 0.153 |
| | BOGD | **30.440 %± 0.991** | 0.025 | 30.760 %± 1.029 | 0.037 | 30.540 %± 0.559 | 0.040 |
| | BOGD++ | 31.080 %± 0.963 | 0.028 | **30.580 %± 0.963** | 0.050 | **30.200 %± 1.054** | 0.062 |
| Budget Size | | B=100 | | B=200 | | B=300 | |
| Dataset | Algorithm | Mistake (%) | Time (s) | Mistakes (%) | Time (s) | Mistakes (%) | Time (s) |
| spambase | RBP | 34.153 %± 0.657 | 0.065 | 32.236 %± 0.241 | 0.132 | 30.585 %± 0.943 | 0.193 |
| | Forgetron | 34.658 %± 0.463 | 0.117 | 32.436 %± 0.715 | 0.231 | 30.785 %± 0.888 | 0.320 |
| | Projectron | 31.841 %± 0.398 | 0.147 | 30.467 %± 4.524 | 0.426 | 29.302 %± 4.831 | 0.847 |
| | Projectron++ | 31.302 %± 0.293 | 0.495 | **28.468 %± 0.702** | 1.752 | 29.359 %± 4.959 | 4.013 |
| | BOGD | 31.158 %± 0.500 | 0.089 | 29.572 %± 0.437 | 0.180 | 28.472 %± 0.785 | 0.267 |
| | BOGD++ | **31.128 %± 0.357** | 0.096 | 28.732 %± 0.929 | 0.193 | **28.329 %± 0.280** | 0.282 |
| Budget Size | | B=200 | | B=400 | | B=600 | |
| Dataset | Algorithm | Mistake (%) | Time (s) | Mistakes (%) | Time (s) | Mistakes (%) | Time (s) |
| w8a | RBP | 4.793 %± 0.069 | 2.566 | 4.200 %± 0.072 | 4.868 | 3.906 %± 0.099 | 7.134 |
| | Forgetron | 4.868 %± 0.073 | 2.656 | 4.203 %± 0.024 | 5.976 | 3.888 %± 0.037 | 8.206 |
| | Projectron | 3.103 %± 0.019 | 3.044 | 3.214 %± 0.087 | 7.748 | 3.202 %± 0.061 | 14.546 |
| | Projectron++ | 3.103 %± 0.014 | 3.670 | 2.934 %± 0.078 | 12.398 | 2.783 %± 0.046 | 23.728 |
| | BOGD | 3.038 %± 0.016 | 2.710 | 3.627 %± 0.108 | 6.108 | 3.339 %± 0.141 | 8.974 |
| | BOGD++ | **3.037 %± 0.007** | 2.938 | **2.724 %± 0.144** | 6.850 | **2.701 %± 0.047** | 9.478 |
| Budget Size | | B=500 | | B=1000 | | B=1500 | |
| Dataset | Algorithm | Mistake (%) | Time (s) | Mistakes (%) | Time (s) | Mistakes (%) | Time (s) |
| magic04 | RBP | 31.682 %± 0.363 | 0.740 | 30.268 %± 0.341 | 1.557 | 29.402 %± 0.325 | 2.421 |
| | Forgetron | 31.891 %± 0.243 | 0.980 | 30.521 %± 0.288 | 1.968 | 29.831 %± 0.400 | 2.905 |
| | Projectron | 28.076 %± 0.590 | 8.280 | 27.361 %± 0.424 | 28.419 | 27.089 %± 0.339 | 61.797 |
| | Projectron++ | 28.073 %± 0.552 | 50.108 | 27.357 %± 0.421 | 173.576 | 27.089 %± 0.310 | 366.590 |
| | BOGD | 28.019 %± 0.450 | 0.803 | 25.724 %± 0.477 | 1.697 | 24.957 %± 0.348 | 2.723 |
| | BOGD++ | **27.255 %± 0.714** | 1.009 | **25.211 %± 0.422** | 2.079 | **24.368 %± 0.423** | 3.312 |
| Budget Size | | B=500 | | B=1000 | | B=2000 | |
| Dataset | Algorithm | Mistake (%) | Time (s) | Mistakes (%) | Time (s) | Mistakes (%) | Time (s) |
| ijcnn1 | RBP | 15.621 %± 0.162 | 7.654 | 15.401 %± 0.173 | 18.463 | 15.270 %± 0.139 | 34.046 |
| | Forgetron | 16.723 %± 0.541 | 9.657 | 16.006 %± 0.308 | 22.411 | 15.273 %± 0.134 | 41.723 |
| | Projectron | 16.103 %± 0.686 | 32.490 | 15.103 %± 0.666 | 75.250 | 13.203 %± 0.581 | 219.310 |
| | Projectron++ | 15.373 %± 0.037 | 35.070 | 14.074 %± 0.042 | 109.270 | 12.223 %± 0.258 | 363.520 |
| | BOGD | 16.505 %± 0.652 | 8.441 | 16.176 %± 0.554 | 19.875 | 13.614 %± 0.320 | 38.787 |
| | BOGD++ | **15.225 %± 0.488** | 8.833 | **13.238 %± 0.550** | 20.321 | **12.117 %± 0.238** | 39.086 |
| Budget Size | | B=500 | | B=1000 | | B=2000 | |
| Dataset | Algorithm | Mistake (%) | Time (s) | Mistakes (%) | Time (s) | Mistakes (%) | Time (s) |
| codrna | RBP | 17.130 %± 0.078 | 11.519 | 16.139 %± 0.046 | 26.736 | 15.532 %± 0.051 | 58.715 |
| | Forgetron | 16.773 %± 0.069 | 13.275 | 15.962 %± 0.120 | 30.298 | 15.316 %± 0.052 | 65.292 |
| | Projectron | 16.883 %± 0.606 | 58.718 | 16.375 %± 0.666 | 312.179 | 15.333 %± 0.540 | 1287.570 |
| | Projectron++ | 15.967 %± 0.721 | 208.015 | 15.025 %± 0.743 | 851.189 | 14.636 %± 0.815 | 1926.070 |
| | BOGD | 18.504 %± 0.236 | 11.601 | 18.465 %± 0.225 | 27.471 | 15.274 %± 0.660 | 56.439 |
| | BOGD++ | **15.634 %± 0.603** | 12.313 | **14.418 %± 0.206** | 28.552 | **13.439 %± 0.220** | 61.181 |

Table 3. Evaluation of several budgeted algorithms with different budgets on six data sets.

Finally, for the comparison of running time cost, the Projectron algorithms are the least efficient algorithms among all the budget online learning algorithms, mostly due to their costly projection step. For example, on the largest dataset "codrna" with the budget B=2000, Projectron++ on average took more than half an hour to run one experiment. For the proposed algorithms, the time costs of both BOGD and BOGD++ are in general comparable to those of RBP and Forgetron, and are significantly more efficient than those of Projectron algorithms. For example, on dataset "codrna" with the budget B=1000, BOGD++ is about 30 times faster than Projectron++, and is about 60 times faster than the original non-budget OGD algorithm. For the two proposed algorithms themselves, BOGD++ is slightly more time consuming than BOGD, due to the additional cost of computing the distribution $p_t$ towards non-uniform sampling.

## 4. Conclusions

This paper presented a novel framework of bounded online gradient descent (BOGD) for scalable kernel-based online learning which requires the number of support vectors to be smaller than a predefined budget. The basic idea of maintaining the budget size is to remove one randomly selected support vector whenever the support vector size overflows. In particular, we proposed two efficient BOGD algorithms: (i) BOGD by randomly discarding one support vector using uniform sampling, and (ii) BOGD++ using non-uniform sampling. We conducted extensive empirical studies by comparing with several state-of-the-art algorithms, in which our results showed that the proposed algorithms achieve the promising performance in terms of both classification efficacy and computational efficiency. Future work will exploit different sampling strategies and extend our work to multi-class budget kernel-based online learning.

## Acknowledgments

This work was in part supported by MOE tier 1 grant (RG33/11), Microsoft Research grant (M4060936), National Science Foundation (IIS-0643494), and Office of Navy Research (ONR Award N00014-09-1-0663 and N00014-12-1-0431).